\pdfoutput=1
\documentclass[10pt,twocolumn,letterpaper]{article}
\usepackage{iccv}
\usepackage{times}
\usepackage{epsfig}
\usepackage{graphicx}
\usepackage{amsmath}
\usepackage{amssymb}
\usepackage{booktabs}
\usepackage{soul}
\usepackage{multirow}
\usepackage{multicol, caption}
\usepackage{paralist, tabularx}
\usepackage[pagebackref=true,breaklinks=true,letterpaper=true,colorlinks,bookmarks=false]{hyperref}

\usepackage[capitalize]{cleveref}
\crefname{section}{Sec.}{Secs.}
\Crefname{section}{Section}{Sections}
\Crefname{table}{Table}{Tables}
\crefname{table}{Tab.}{Tabs.}

\iccvfinalcopy

\begin{document}
\title{
PixelRNN: In-pixel Recurrent Neural Networks for End-to-end-optimized Perception with Neural Sensors}
\author{Haley M. So$^1$ $\quad$ Laurie Bose$^2$ $\quad$ Piotr Dudek$^2$ $\quad$ Gordon Wetzstein$^1$\\
$^1$Stanford University $\quad$ $^2$The University of Manchester\\
{\tt\small \href{https://www.computationalimaging.org/publications/pixelrnn/}{computationalimaging.org/publications/pixelrnn/} }
}
\maketitle
\begin{abstract}
Conventional image sensors digitize high-resolution images at fast frame rates, producing a large amount of data that needs to be transmitted off the sensor for further processing. This is challenging for perception systems operating on edge devices, because communication is power inefficient and induces latency. Fueled by innovations in stacked image sensor fabrication, emerging sensor--processors offer programmability and minimal processing capabilities directly on the sensor. We exploit these capabilities by developing an efficient recurrent neural network architecture, PixelRNN, that encodes spatio-temporal features on the sensor using purely binary operations. PixelRNN reduces the amount of data to be transmitted off the sensor by a factor of $64 \times$ compared to conventional systems while offering competitive accuracy for hand gesture recognition and lip reading tasks. We experimentally validate PixelRNN using a prototype implementation on the SCAMP-5 sensor--processor platform.
\end{abstract}
\section{Introduction}
\label{sec:intro}

Increasingly, cameras on edge devices are being used for enabling computer vision perception tasks rather than for capturing images that look beautiful to the human eye. Applications include various tasks in virtual and augmented reality displays, wearable computing systems, drones, robotics, and the internet of things, among many others. For such edge devices, low-power operation is crucial, making it challenging to deploy large neural network architectures which traditionally leverage modern graphics processing units for inference.

A plethora of approaches have been developed in the ``TinyML'' community to address these challenges. Broadly speaking, these efforts focus on developing smaller~\cite{Howard:mobilenets} or more efficient network architectures, often by pruning or quantizing larger models~\cite{ChengQuant}. Platforms like TensorFlow Lite Micro enable application developers to deploy their models directly to power-efficient microcontrollers which process data closer to the sensor. Specialized artificial intelligence (AI) accelerators, such as Movidius' Myriad vision processing unit, further reduce the power consumption. While these approaches can optimize the processing component of a perception system, they do not reduce the large amount of digitized sensor data that needs to be transmitted to the processor in the first place, via power-hungry interfaces such as MIPI-CSI, and stored in the memory. This omission is highly significant as data transmission and memory access are among the biggest power sinks in imaging systems~\cite{GomezPower}. This raises the question of how to design perception systems where sensing, data communication, and processing components are optimized end to end.

Efficient perception systems could be designed such that important task-specific image and video features are encoded directly on the imaging sensor using in-pixel processing, resulting in the sensor's output being significantly reduced to only these sparse features. This form of in-pixel feature encoding mechanism could significantly reduce the required bandwidth, thus reducing power consumption of data communication, memory management, and downstream processing. Event sensors~\cite{Gallego2022EventBasedVA} and emerging focal-plane sensor--processors~\cite{Zarandy:2011} are promising hardware platforms for such perception systems because they can directly extract either temporal information or spatial features, respectively, on the sensor. These features can be transmitted off the sensor using low-power parallel communication interfaces supporting low bandwidths.

Our work is motivated by the limitations of existing feature extraction methods demonstrated on these emerging sensor platforms. Rather than extracting simple temporal gradients~\cite{Gallego2022EventBasedVA} or spatial-only features via convolutional neural networks (CNNs)~\cite{boseECCV2020,cameraThatCNNs2019}, we propose in-pixel recurrent neural networks (RNNs) that efficiently extract spatio-temporal features on sensor--processors for bandwidth-efficient perception systems. RNNs are state-of-the-art network architectures for processing sequences, such as video in computer vision tasks~\cite{RNNreview}. Inspired by the emerging paradigm of neural sensors~\cite{Martel2020NeuralSensors}, our in-pixel RNN framework, dubbed PixelRNN, comprises a light-weight in-pixel spatio-temporal feature encoder. This in-pixel network is jointly optimized with a task-specific downstream network. We demonstrate that our architecture outperforms event sensors and CNN-based sensor--processors on perception tasks, including hand gesture recognition and lip reading, while drastically reducing the required bandwidth compared to any traditional sensor based approaches. Moreover, we demonstrate that PixelRNN offers better performance and lower memory requirements than larger RNN architectures in the low-precision settings of in-pixel processing.

Our work's contributions include 
\begin{compactitem}
    \item the design and implementation of in-pixel recurrent neural networks for sensor--processors, enabling bandwidth-efficient perception on edge devices;
    \item the demonstration that our on-sensor spatio-temporal feature encoding maintains high performance while significantly reducing sensor-to-processor communication bandwidth with several tasks, including hand gesture recognition and lip reading;
    \item the experimental demonstration of the benefits of in-pixel RNNs using a prototype implementation on the SCAMP-5 sensor--processor. 
\end{compactitem}

\section{Related Work}
\label{sec:related}
Performing feature extraction on power and memory constrained computing systems requires the union of multiple fields: machine learning, specialized hardware, and network compression techniques.

\paragraph{Machine Learning on the Edge.} Edge computing devices are often subject to severe power and memory constraints, leading to various avenues of research and development. On the hardware side, approaches include custom application-specific integrated circuits (ASICs), field-programmable gate arrays (FPGAs), or other energy efficient AI accelerators. However, this does not address the issue of data transmission from imaging sensors, which is one of the main sources of power consumption~\cite{GomezPower}. To circumvent the memory constraints, network compression techniques are introduced. They fall into roughly five categories \cite{ChengQuant}: 1. parameter reduction by pruning redundancy~\cite{Blalock2020WhatIT, Srinivas2015DatafreePP, Huang2018DataDrivenSS, LessIsMore}; 2. low-rank parameter factorization \cite{Denton, Jaderberg, Tai2016ConvolutionalNN}; 3. carefully designing structured convolutional filters  \cite{Dieleman,Szegedy,SqueezeDet}; 4. creating smaller models \cite{Hinton2015DistillingTK,DoDeepNets,Bucila2006ModelC}; 5. parameter quantization \cite{binaryConnect,xnor,BNN,Tang,birealnet,DoRe,abcNet}. In this work, we move compute directly onto the sensor and also apply ideas and techniques mentioned above.

\paragraph{Beyond Frame-based Sensing.} %
Event-based cameras have been gaining popularity \cite{Gallego2022EventBasedVA} as the readout is asynchronous and often sparse, triggered by pixel value changes above a certain threshold. However, these sensors are not programmable and do data compression with a simple fixed function. Another emerging class of sensors include focal plane sensor--processors, also known as pixel processor arrays. Along with supporting traditional sensing capabilities, these sensors have a processing element embedded into each pixel. While conventional vision systems have separate hardware for sensing and computing, sensor--processors perform both tasks ``in pixel,'' enabling efficient, low-latency and low-power computation. Recently, sensor--processors with some programmability have emerged~\cite{carey2013100,Miao,Lopich,Rodriguez,Zhang_chip,Berni}. Further advances in 3D fabrication techniques, including wafer-level hybrid bonding and stacked CMOS image sensors, set the stage for rapid development of increasingly more capable programmable sensors. 

\paragraph{In-pixel Perception.} %
In the past few years, there has been a surge of advances in neural networks for vision tasks as well as an increasing desire to perform tasks on constrained mobile and wearable computing systems. Sensor--processors are a natural fit for such systems as they can perform sophisticated visual computational tasks at a significantly lower power than traditional hardware. Some early chips \cite{ACE4k, ACE16k} were based on implementing convolution kernels in a recurrent dynamical ``Cellular Neural Network'' model \cite{Chua1988, Roska1993}. In 2019, Bose et al. created ``A Camera that CNNs'' -- one of the first works to implement a deep convolutional neural network on the sensor~\cite{cameraThatCNNs2019}. 
Since then, there have been a number of other works in CNNs on programmable sensors~\cite{cain,auke,analognet,guillard_2019,boseECCV2020,Liu2020HighspeedLC,Liu_2022_CVPR,liu_neuro}. These works extract features in the spatial domain, but miss a huge opportunity in failing exploit temporal information. Purely CNN based approaches do not utilize or capitalize on the temporal redundancy or information of the sequence of frames. Our work introduces light-weight extraction of spatio-temporal features, better utilizing the structure of the visual data, all while maintaining low bandwidth and high accuracy.

\section{In-pixel Recurrent Neural Networks}
\label{sec:method}
\begin{figure*}[t]
    \centering
    \includegraphics[clip, trim=0cm 9cm 0.25cm 2.5cm, width=1\linewidth]{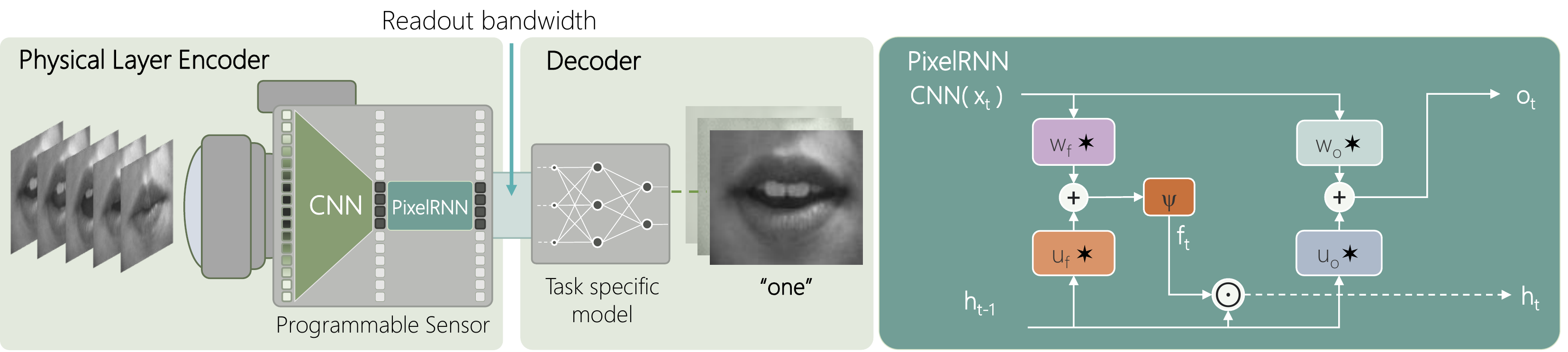}
    \caption{The perception pipeline of PixelRNN can be broken down into an on-sensor encoder and a task-specific decoder. On the left is the camera equipped with a sensor--processor, which offers processing and memory at the pixel level. The captured light is directly processed by a CNN that extracts spatial features, which are further processed by a convolutional recurrent neural network with built-in memory and temporal feature extraction. Here we show our PixelRNN variant on the right, $\star$ being the convolution operator, $\odot$ element-wise multiplication, and $\psi$ a nonlinear function. Instead of sending out full $256\times 256$ values at every time step, our encoder compresses by $64\times$. While we show this pipeline for a lip reading task, the decoder can be designed for any perception task.  
    }
    \label{fig:Pipeline2}
\end{figure*}

Emerging sensor--processors with in-pixel processing enable the development of end-to-end-optimized on-sensor and downstream networks off-sensor. In this section, we describe a new on-sensor recurrent spatio-temporal feature encoder that significantly improves upon existing temporal- or spatial-only feature encoders for video processing, as shown in the next section. The proposed pipeline is illustrated in Figure~\ref{fig:Pipeline2}.

\subsection{In-Pixel CNN-based Feature Encoder}

Convolutional neural networks are among the most common network architectures in computer vision. They are written as
\begin{align}
    \textrm{ \sc{CNN} } \left( \mathbf{x} \right) & =  \left( \phi_{n-1} \circ \phi_{n-2} \circ \ldots \circ \phi_{0} \right) \left( \mathbf{x} \right) , \nonumber \\
    \phi_i  = \mathbf{x}_i & \mapsto \psi \left( \mathbf{w}_i * \mathbf{x}_i + \mathbf{b}_i \right),
\end{align}
where $\mathbf{w}_i * \mathbf{x}_i : \mathbb{N}^{N_i \times M_i \times C_i} \mapsto \mathbb{N}^{N_{i+1}\times M_{i+1} \times C_{i+1}}$ describes the multi-channel convolution of CNN layer $i$ and $\mathbf{b}_i$ is a vector containing bias values. Here, the input image $\mathbf{x}_i$ has $C_i$ channels and a resolution of $N_i \times M_i$ pixel and the output of layer $i$ is further processed by the nonlinear activation function $\psi$.

The SCAMP-5 system used in this work lacks native multiplication operations at the pixel level. Due to this limitation, works storing network weights $\mathbf{w}_i$ in pixel typically restrict themselves to using binary, $\left\{-1,1\right\}$, or ternary $\left\{-1,0,1\right\}$ values. This reduces all multiplications to sums or differences, which are highly efficient native operations.

\subsection{In-pixel Spatio-temporal Feature Encoding}

Recurrent neural networks (RNNs) are state-of-the-art network architectures for video processing. Whereas a CNN only considers each image in isolation, an RNN extracts spatio-temporal features to process video sequences more effectively. Network architectures for sensor--processors must satisfy two key criteria. First, they should be small and use low-precision weights. Second, they should comprise largely of local operations as the processors embedded within each pixel can only communicate with their direct neighbors (e.g.,~\cite{carey2013100}). 

To satisfy these unique constraints, we devise an RNN architecture that combines ideas from convolutional gated recurrent units (GRUs)~\cite{ballas2015delving} and minimal gated units~\cite{minimalGRU}. The resulting simple, yet effective PixelRNN architecture, is written as 
\begin{align}
    \mathbf{f}_t &= \psi_f \left ( \mathbf{w}_f \! * {\text{CNN} \left(\mathbf{x}_t\right) }+ \mathbf{u}_f*\mathbf{h}_{t-1} \right ),\\
    \mathbf{h}_t &= \mathbf{f}_t \odot \mathbf{h}_{t-1}, \label{eq:forget_gate}\\
    \mathbf{o}_t &= \psi_o \left (\mathbf{w}_o * {\text{CNN} \left(\mathbf{x}_t\right) } + \mathbf{u}_o * \mathbf{h}_{t-1} \right ), \label{eq:pixelrnn:output}
\end{align}
where $\mathbf{w}_f$, $\mathbf{u}_f$, $\mathbf{w}_o$, $\mathbf{u}_o$ are small convolution kernels and $\psi_f$ is either the sign (when working with binary constraints) or the sigmoid function (when working with full precision). We include an optional nonlinear activation function $\psi_o$ and an output layer $\mathbf{o}_t$ representing the values that are actually transmitted off sensor to the downstream network running on a separate processor. For low-bandwidth operation, the output layer $o$ is only computed, and values transmitted off the sensor, every $K$ frames. The output layer can optionally be omitted, in which case the hidden state $\mathbf{h}_t$ is streamed off the sensor every $K$ frames.

PixelRNN uses what is commonly known as a ``forget gate'', $\mathbf{f}_t$, and a hidden state $\mathbf{h}_t$, which are updated at each time step $t$ from the input $\mathbf{x}_t$. RNNs use forget gates to implement a ``memory'' mechanism that discards redundant spatial features over time. PixelRNN's forget gate is also motivated by this intuition, but our update mechanism in Eq.~\ref{eq:forget_gate} is tailored to working with binary constraints using values $\left\{-1,1\right\}$. In this case, Eq.~\ref{eq:forget_gate} flips the sign of $\mathbf{f}_t$ in an element-wise manner rather than decaying over time. This mechanism works very well in practice when $\mathbf{h}_t$ is re-initialized to all ones every 16 time steps.

PixelRNN's architecture includes alternative spatial- or temporal-only feature extractors as special cases. For example, it is intuitive to see that it models a conventional CNN by omitting the recurrent units. 
We specifically write out the output gate in our definition to make it intuitive how PixelRNN also approximates a difference camera as a special case, which effectively implement a temporal-only feature encoder. In this case, $\mathbf{h}_t=\mathbf{x}_t, \text{w}_o = 1, \text{u}_o=-1$ and \\
$\psi_o \left(x \right) = \left\{ \footnotesize{\begin{array}{rl}
        -1 & \text{for } \textrm{x} < - \delta\\
        0 & \text{for } -\delta \leq \textrm{x} \leq \delta\\
        1 & \text{for } \delta < \textrm{x}
        \end{array} } \right. ,
$
for some threshold $\delta$. The image formation model of event cameras~\cite{Gallego2022EventBasedVA} is asynchronous and a difference camera represents only a crude approximation, but it serves as a pedagogically useful temporal-only encoder in the context of this discussion.

\subsection{Learning Quantized In-pixel Parameters}

PixelRNN uses binary weights to reduce all multiplications to sums. To learn these parameters efficiently, we parameterize each of these values $w$ using a continuous value $\widetilde{w}$ and a quantization function $q$ such that
\begin{equation}
    w = q \left( \widetilde{w} \right), \quad q: \mathbb{R} \rightarrow \mathcal{Q}.
\end{equation}
Here, $q$ maps a continuous value to the closest discrete value in the feasible set $\mathcal{Q}$, i.e., $\left\{-1,1\right\}$. 

One can employ surrogate gradient methods~\cite{bengio2013estimating,zenke2018superspike}, continuous relaxation of categorical variables using Gumbel-Softmax~\cite{jang2016categorical,maddison2016concrete}, or other approaches to approximately differentiate through $q$. 
For the binary weights we use, $w = q(\tilde{w}) = \text{sign}(\tilde{w})$, and we found approximating the gradient of the sign function with the derivative of $\textrm{tanh}(mx)$ produced very good results:
\begin{equation}
\frac{\partial q}{\partial\tilde{w}}  \approx m\cdot (1- \tanh^2(m\tilde{w}))
\end{equation}
where $m>0$ controls the steepness of the $\tanh$ function, which is used as a differentiable proxy for $q \approx \tanh(m\tilde{w})$ in the backward pass. The larger $m$ is, the more it resembles the sign function and the more the gradient resembles the delta function.

\subsection{Implementation Details}

We implement our per-frame CNN feature encoder, testing both 1 or 2 layers and process images at a resolution of $64 \times 64$ pixels by downsampling the raw sensor images before feeding them into the CNN. In all experiments, our PixelRNN transmits data off the sensor every 16 frames. Thus, we achieve a reduction in bandwidth by a factor of 64$\times$ compared to the raw data.
In all of our experiments, we set the function $\psi_o$ to the identity function.

Additional implementation details are found in the supplement and source code will be made public for enhanced reproducibility.

\section{Experiments}
\label{sec:results}
\paragraph{Evaluating Feature Encoders.}

As discussed in the previous section, RNNs require a CNN-based feature encoder as part of their architecture. In-pixel CNNs have been described in prior work~\cite{boseECCV2020,Liu2020HighspeedLC,Liu_2022_CVPR}, albeit not in the context of video processing with RNNs. 

Table~\ref{tab:CNN_comparison} summarizes the simulation performance of re-implementations of various CNN architectures on image classification using the MNIST, CIFAR-10 dataset, and hand gesture recognition from individual images.

Bose et al.~\cite{cameraThatCNNs2019, boseECCV2020} describes a 2-layer CNN with ternary weights. The two works have the same architecture but differ drastically in the sensor--processor implementation. Liu et al.~\cite{Liu2020HighspeedLC,Liu_2022_CVPR} describes 1- and 3-layer CNNs with binary weights using a different architecture than Bose while using similar sensor--processor implementations concepts. Our feature encoder is a binary 2-layer variant of Bose et al.'s CNN. Each layer has 16 kernels of size $5 \times 5$ and are followed with a non-linearity and maxpooling of $4 \times 4$. The 16 $16 \times 16$ channels are then concatenated into a single $64 \times 64$ image size to serve as the input to the next convolutional layer or to the PixelRNN. All of these CNNs are roughly on-par with some performing better at some tasks than others. Ours strikes a good balance between accuracy and model size. We do not claim this CNN to be a contribution of our work, but include this brief comparison for completeness.

\begin{table}[]
    \centering
    \scriptsize
    \begin{tabular}{cccccccc}
    \toprule
         Model Name & MNIST&CIFAR-10& Hand & \# Model & Size \\
          & && Gesture & Params  &  (MB)   \\
         \midrule
         Bose~\cite{boseECCV2020, cameraThatCNNs2019} & \underline{95.0}\%&\underline{39.8}\%&43.4\% & 257&  0.05 \\
         Liu~2020~\cite{Liu2020HighspeedLC} & 80.0\%&32.5\%&57.4\% & $258$& 0.03 \\
         Liu~2022~\cite{Liu_2022_CVPR} & \textbf{95.1}\%&32.6\%&\underline{60.2}\% & $2,374$ & 0.30 \\
         Our CNN & 90.9\%&\textbf{43.1}\%&\textbf{68.1}\% & $802$ & 0.10  \\
        \bottomrule
    \end{tabular}
    \caption{\textbf{Comparing CNN Feature Encoders.} We simulated different in-pixel CNNs using image classification on MNIST, CIFAR-10, and the Cambridge hand gesture recognition based on different implementations. All CNN architectures perform roughly on par with our binary 2-layer CNN encoder striking a good balance between accuracy and model size. The model size is computed by multiplying the number of model parameters by the quantization of the values.
    }
    \label{tab:CNN_comparison}
\end{table}

\begin{figure*}[t!]
    \centering
    \includegraphics[clip, trim=0cm 0cm 1cm 0cm, width=\columnwidth]{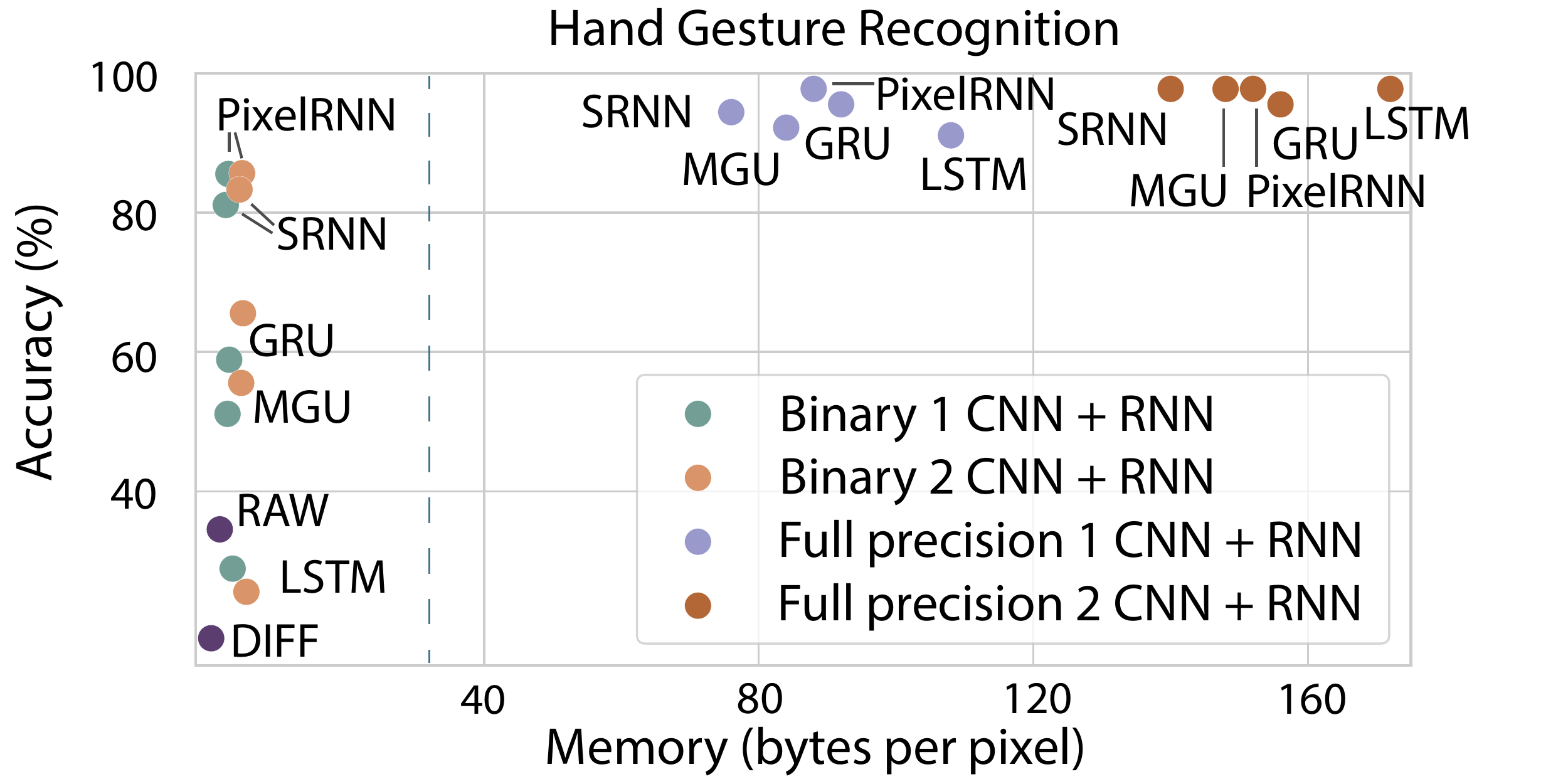}
    \hspace{1em}
    \includegraphics[clip, trim=0.5cm 0cm 1cm 0cm, width=\columnwidth]{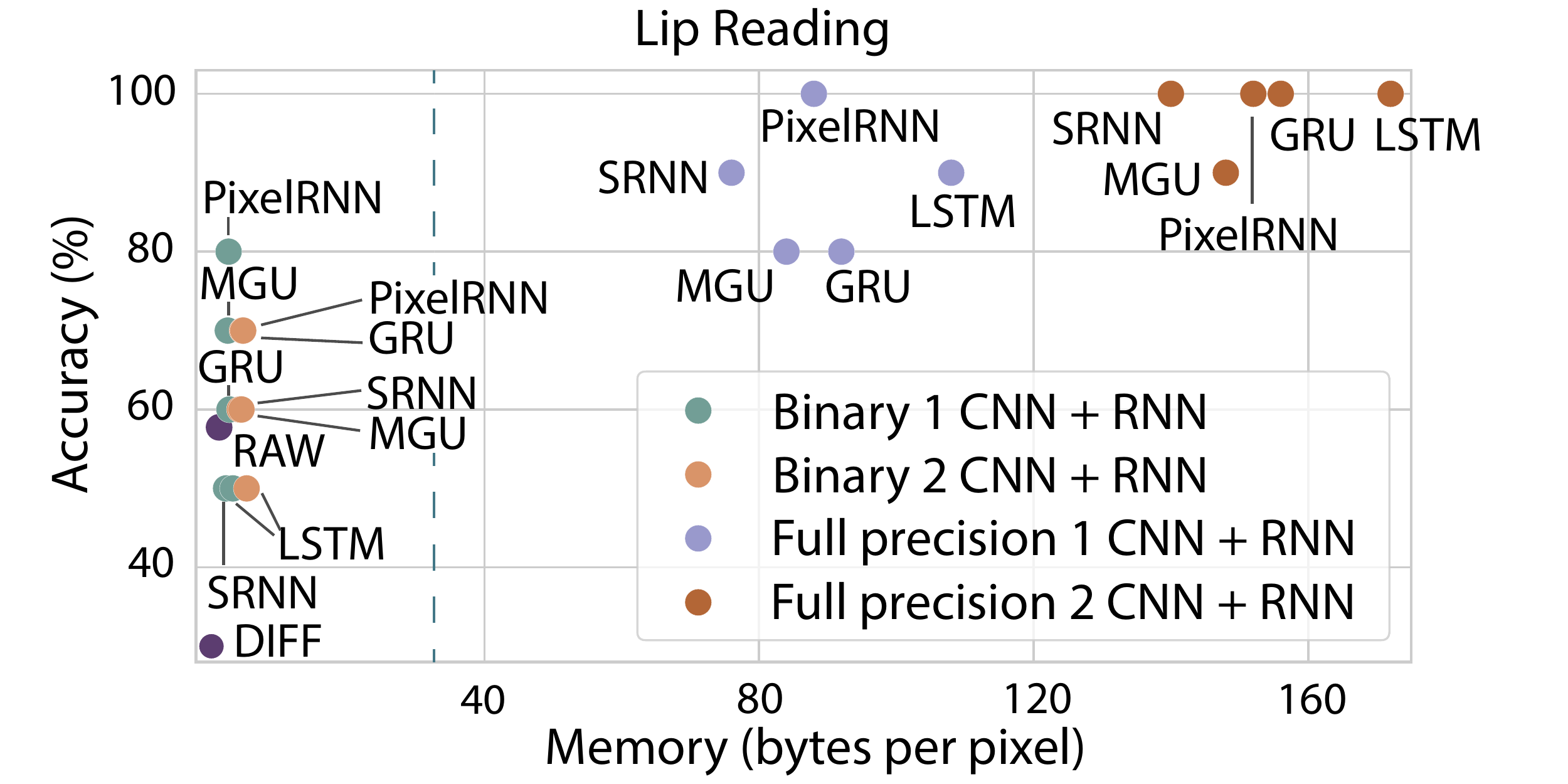}
    \caption{{\bf Network architecture comparison.}  
    We compare baselines, including a RAW and difference camera (DIFF) as well as several RNN architectures, each with 1- and 2-layer CNN encoders and binary or full 32 bit floating point precision. PixelRNN offers the best performance for the lowest memory footprint, especially when used with binary weights. The dashed vertical line indicates the available memory on our hardware platform, SCAMP-5, showing that low-precision network architectures are the only feasible option in practice. 
    }
    \label{fig:accuracyplots}
\end{figure*}

\paragraph{Baseline Architectures.}

We use several baselines for our analysis. The RAW camera mode simply outputs the input frame at every time step and represents the naive imaging approach. The simulated difference camera represents a simple temporal-only feature encoder. 
We also include several RNN architectures, including long short-term memory (LSTM), gated recurrent unit (GRU), minimal gated unit (MGU), a simple RNN (SRNN), and our PixelRNN. Moreover, we evaluated each of the RNN architectures using 1-layer and 2-layer CNN feature encoders. The output of all RNNs is read from the sensor only once every 16 time steps. All baselines represent options for in-pixel processing and their respective output is streamed off the sensor. In all cases, a single fully-connected network layer processes this output on a downstream processor to compute the final classification scores. This fully-connected layer is trained end to end for each of the baselines. While this fully-connected layer is simple, it is effective and could be customized for a specific task. 

Additional details on these baselines, including formulas and training details, are listed in the supplement. Table~\ref{tab:modelcomparison} shows an overview of these, listing the number of model parameters and the readout bandwidth for each of them. 

\begin{table}[]
    \centering    
    \small
    \begin{tabular}{ccccc}
    \toprule
         \multicolumn{2}{c}{Model Name} &  \begin{tabular}{@{}c@{}}\# Model \\ Parameters\end{tabular} & \begin{tabular}{@{}c@{}}Readout  \\ Bandwidth\end{tabular}   \\
         \midrule
         \multicolumn{2}{c}{RAW} &0&  1,048,576 \\
         \multicolumn{2}{c}{Difference Camera}        &0 & 65,536\\ 
         \midrule 
         \multirow{5}{*}{\rotatebox{90}{1-layer CNN}} & SRNN   & 451  &  4,096  \\
         & LSTM   & 601  & 4,096 \\
         &  GRU   & 551 &  4,096  \\
         &  MGU   & 501 &  4,096  \\
         & PixelRNN & 501  &   4,096  \\ 
         \midrule
         \multirow{5}{*}{\rotatebox{90}{2-layer CNN}}&  SRNN   & 852  &  4,096  \\
         &  LSTM   & 1,002  &  4,096  \\
         & GRU   & 952 &  4,096  \\
         & MGU   & 902  &  4,096  \\
         & PixelRNN & 902 &   4,096  \\
         \bottomrule
    \end{tabular}
    \caption{{\bf Overview of Baselines.} We list the number of model parameters and the readout bandwidth (in values) for each baseline. The RAW and difference camera modes stream data off the sensor at every frame. All RNNs only stream data off the sensor once every 16 frames, providing a significant benefit in bandwidth.}
    \label{tab:modelcomparison}
\end{table}
\paragraph{Datasets.}

For the hand gesture recognition task, we use the \href{https://labicvl.github.io/ges_db.htm}{Cambridge Hand Gesture Recognition} dataset. This dataset consists of 900 video clips of 9 gesture classes; each class contains 100 videos. For the lip reading task, we use the \href{https://inc.ucsd.edu/mplab/36/}{Tulips1} dataset. This dataset is a small Audiovisual database of 12 subjects saying the first 4 digits in English; it was introduced in~\cite{TulipsDataset}.

\paragraph{Accuracy vs. Memory.}

In Figure~\ref{fig:accuracyplots}, we evaluate the accuracy of several baseline architectures on two tasks: hand gesture recognition (left) and lip reading (right). We compare the baselines described above, each with 1- and 2-layer CNN encoders and binary or full 32-bit floating point precision. For the full-precision networks, PixelRNN achieves an accuracy comparable to the best models, but it provides one of the lowest memory footprints. Comparing the networks with binary weights, PixelRNN also offers the best accuracy with a memory footprint comparable to the next best method. Surprisingly, larger architectures, such as GRUs and LSTMs, do not perform well when used with binary weights. This can be explained by the increasing difficulty of reliably training increasingly large networks with binary parameter constraints. Leaner networks, such as SRNN and PixelRNN, can be trained more robustly and reliably in these settings. Note that the memory plotted on the x-axis represents all intermediate features, per pixel, that need to be stored during a single forward pass through the RNN. We do not count the network parameters in this plot, because they do not dominate the memory requirements and can be shared among the pixels.

\paragraph{Constraints of the Experimental Platform.}
Our hardware platform, SCAMP-5, provides an equivalent to 32 bytes of memory per pixel for storing intermediate features. This limit is illustrated as dashed vertical lines in Figure~\ref{fig:accuracyplots}, indicating that only low-precision RNN networks are feasible on this platform. The available memory is computed as follows. SCAMP-5 has 6 analog registers per pixel, each we assume is equivalent to 1 byte: 2 registers store the model weights, 2 are reserved for computation, leaving 2 bytes per pixels for the intermediate features. SCAMP-5 consists of an array of $256 \times 256$ pixel processors, however our approach operates on a smaller effective image size of $64 \times 64$. This allows us to consider a single ``pixel'' to comprise of a block of $4\times4$ pixel elements, increasing the effective bytes per pixel to 32 bytes.

\paragraph{Accuracy vs. Bandwidth.}

\begin{figure}[t]
    \centering
    \includegraphics[clip, trim=1.1cm 0cm 1.5cm 0cm, width=\linewidth]{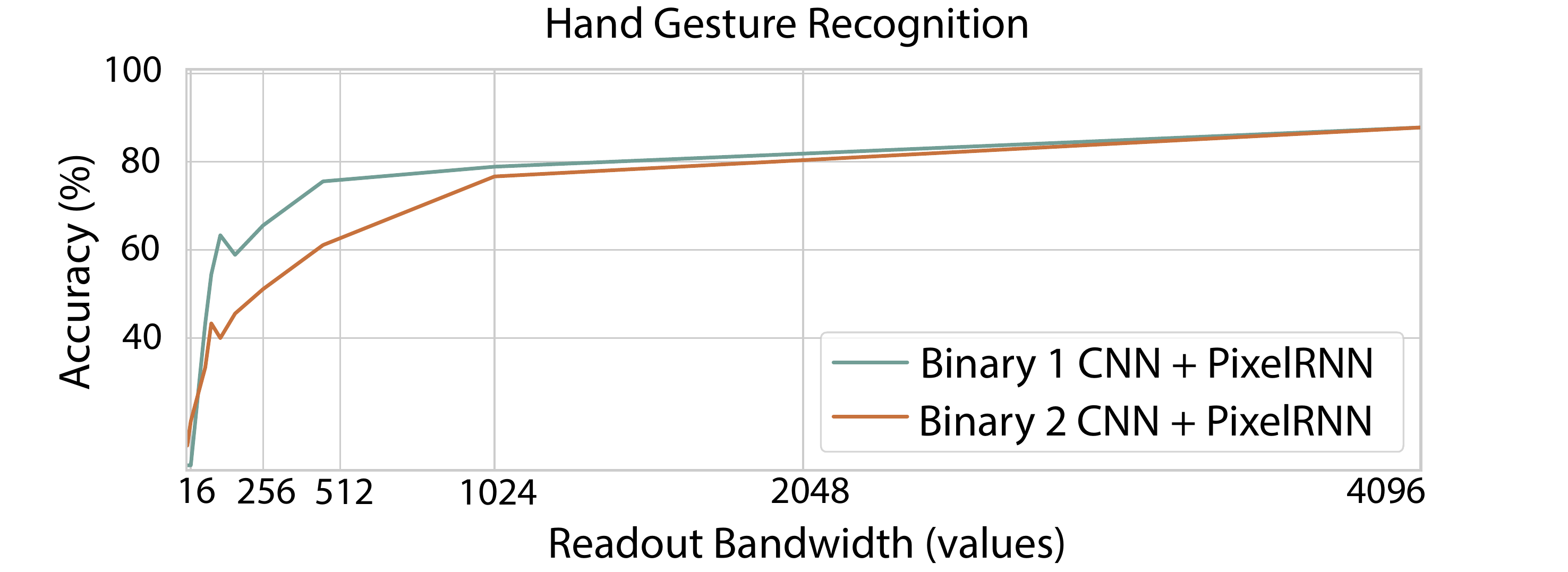}
    \includegraphics[clip, trim=1cm 0cm 1.5cm 0cm, width=\linewidth]{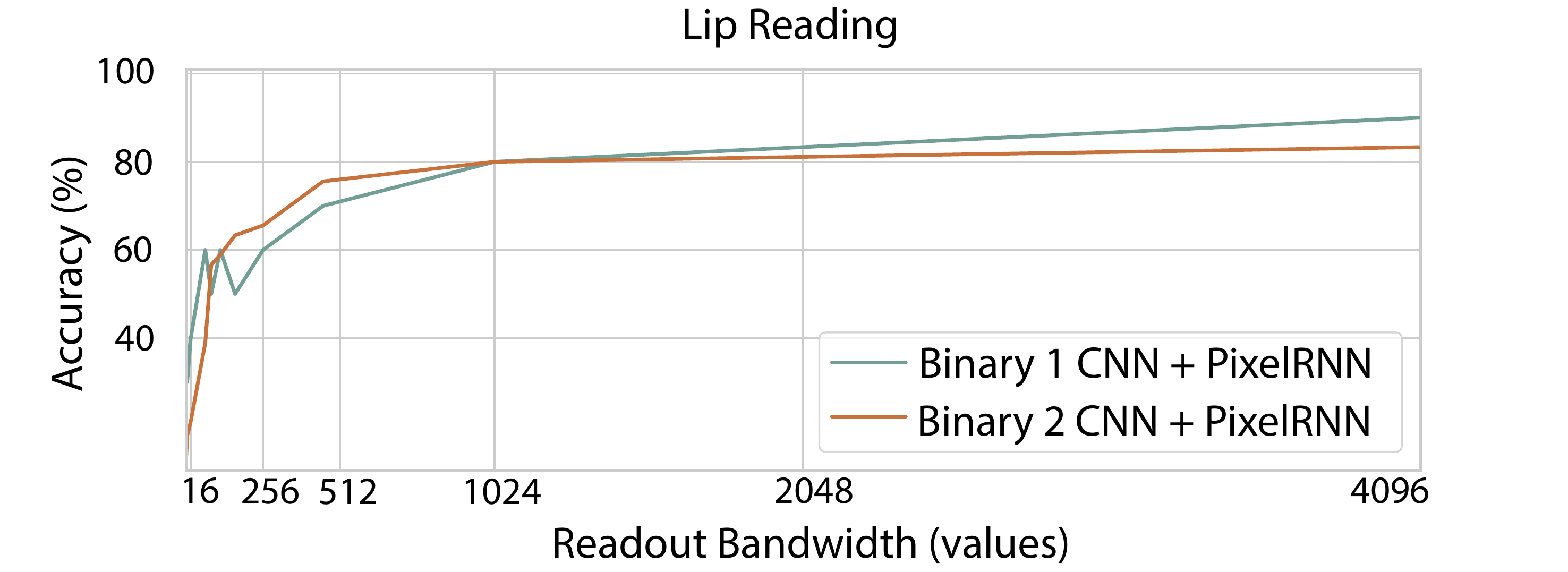}
    \caption{{\bf Bandwidth Analysis.} We can control the bandwidth of data read off the sensor using increasingly larger max-pooling layers before inputting the intensity images to PixelRNN at the cost of decreased accuracy.} 
    \label{fig:bandwidth_study}
\end{figure}

We select a readout bandwidth of 4,096 values (every 16 frames) based on the available bandwidth of our hardware platform, the SCAMP-5 sensor. In Figure~\ref{fig:bandwidth_study} we evaluate the effect of further reducing this bandwidth on the accuracy for the PixelRNN architecture. Bandwidth is controlled using a max-pooling layer operating at differing sizes from $1\times1$ through $8\times 8$ and then at multiples of 8 up to $64\times 64$ before inputting the intensity images to PixelRNN. The resulting output bandwidths range between 1 to 4096. We ran each experiment ten times and the best performances of each are plotted for hand gesture recognition and lip reading. We observe that the bandwidth could be further reduced to about 1,000 values every 16 frames without significantly degrading the accuracy on these datasets. However, decreasing the bandwidth beyond this also reduces the accuracy.

\section{Experimental Prototype}
\label{sec:prototype}
\setlength{\parskip}{0pt}
\subsection{Pixel-Level Programmable Sensors}
SCAMP-5~\cite{carey2013100} is one of the emerging programmable sensors representative of the class of focal-plane sensor--processors (FPSP). Unlike conventional image sensors, each of the $256 \times 256$ pixels is equipped with an arithmetic logic unit, 6 local analog and 13 digital memory registers, control and I/O circuitry, and access to certain registers of the four neighboring pixels. SCAMP-5 operates in single-instruction multiple-data (SIMD) mode with capabilities to address individual pixels, patterns of pixels, or the whole array. It features mixed-mode operation execution that allows for low-power compute prior to A/D conversion. Most importantly, SCAMP-5 is programmable.

\subsection{Implementation of PixelRNN on SCAMP-5}
\begin{figure*}[t]
    \centering
    \includegraphics[clip, trim=0cm 10.4cm 0cm 0cm, width=1\linewidth]{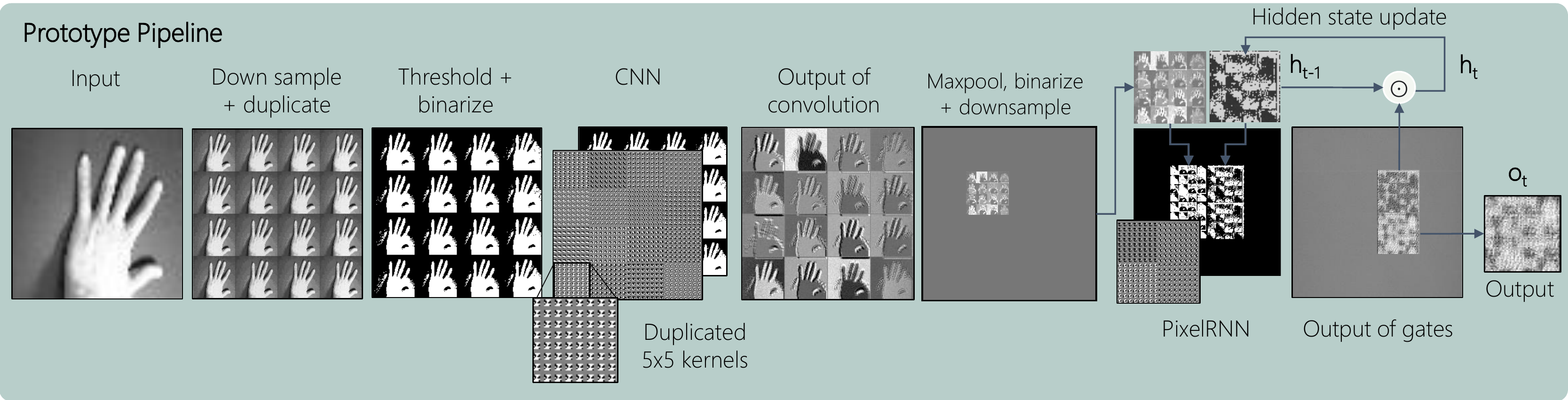}
    \caption{This pipeline shows the sequence of operations from left to right. The input image is downsampled, duplicated, and binarized. Stored convolutional weights perform 16 convolutions, to produce 16 feature maps in the $4 \times 4$ grid of processor elements. A ReLU activation is applied, followed by max-pooling, downsampling, and binarization. This can either be fed to another CNN layer or to the input of the RNN. The RNN takes in the output of the CNN and the previous hidden state $\textbf{h}_{t-1}$. The hidden state $\textbf{h}_t$ is updated every timestep. The output $\textbf{o}_t$ is read out every 16 frames, yielding 64$\times$ decrease in bandwidth. } 
    \label{fig:scampfigurepipeline}
\end{figure*}

\begin{figure*}
    \centering
    \includegraphics[clip, trim=0cm 4.5cm 3cm 0cm, width=0.8\linewidth]{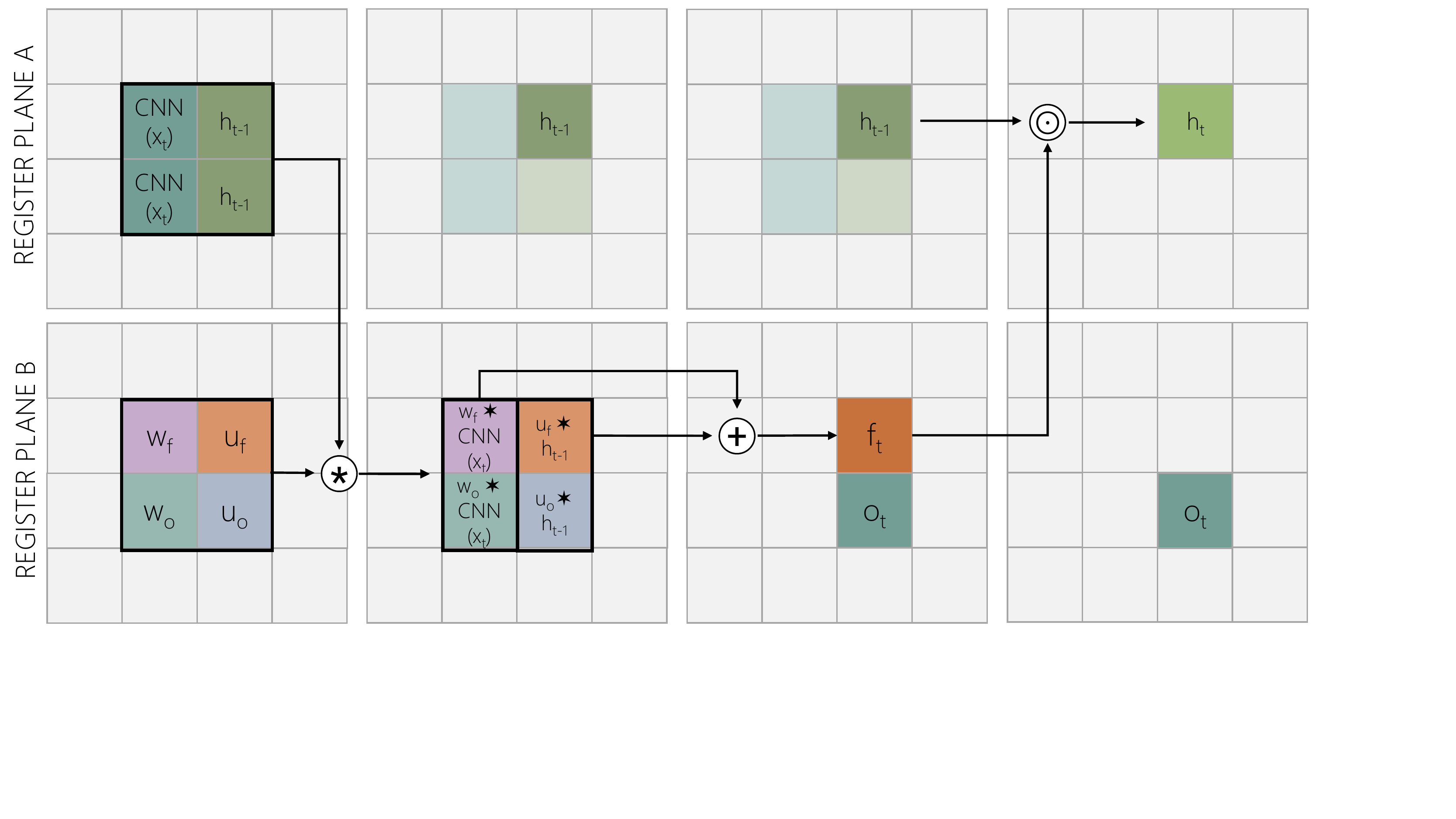}
    \caption{{ To implement the PixelRNN on SCAMP-5, the image plane is split into a $4 \times 4$ grid of processor elements shown above. Two analog register planes are used, Register planes A and B. Above, we show the sequence of operations from left to right. The input from the CNN and the previous hidden state are duplicated in A. These 4 PEs are convolved $*$ with the corresponding gate weights stored in plane B. The resulting convolutions in the second column are then added to compute the output $\textbf{o}_t$ and the forget gate $\textbf{f}_t$. Note that an in-place binarization is applied to $\textbf{f}_t$. The hidden state $\textbf{h}_t$ is updated via an element-wise multiplication $\odot$  of $\textbf{h}_{t-1}$ and $\textbf{f}_t$}.  
    }
    \label{fig:rnnpipeline}
\end{figure*}
{The pipeline for our prototype feature extractor is shown in Figure \ref{fig:scampfigurepipeline}. Because of the memory architecture on SCAMP-5, performing multiple convolutions and different updates of gates and states require us to split the focal plane into 16 parallel processors with a smaller effective image size. The input image is binarized, downsampled to $64 \times 64$, and duplicated out to a $4 \times 4$ grid of parallel processor elements (PE) of size $64 \times 64$. Each PE performs a convolution with a $5 \times 5$ kernel, yielding 16 feature maps per convolutional layer. The 16 $64 \times 64$ features undergo a ReLU activation, maxpooling and binarization to 16 $16 \times 16$. These are then concatenated to create a single $64 \times 64$ input to the RNN or another convolutional layer if desired. This process makes use of the image transformation methods for SCAMP-5 introduced by~\cite{bose2017visual}. Our RNN uses both the output of the CNN and the hidden state to update the hidden state and compute an output every 16 time steps. The RNN gates are calculated via convolution and element-wise multiplication. To suit the SCAMP-5 architecture, we limited operations to addition, XOR, and negation, and trained a binary version of PixelRNN, binarizing weights and features to -1 and 1. Instead of multiplications, we now just need addition and subtraction. }
\paragraph{Memory Allocation.} SCAMP-5's analog and digital registers are limited in number and present different challenges. Analog registers cannot hold values for long periods before decaying. The decay is exacerbated if one moves information from pixel to pixel such as in shifting an image. We found using analog registers with a routine to refresh their content to a set of quantized values inspired by ~\cite{boseECCV2020} helped circumvent some of the challenges. This allowed the storage of binary weights for convolutions and the hidden state for prolonged periods of time. The remaining memory registers were used for performing computations and storing intermediate feature maps. 
\paragraph{Convolution Operation.}
 A single pixel cannot hold all weights of a single kernel, so the weights are spread across a single analog register plane as shown in Figure~\ref{fig:scampfigurepipeline}. To perform a convolution, SCAMP-5 iterates through all 25 weights in the $5 \times 5$ kernel, each time multiplying it with the whole image and adding to a running sum. The image is then shifted, the next weight fills the register plane, and the process continues until the feature is computed. We include a detailed diagram in the supplement and more information can be found in ~\cite{boseECCV2020}.

\paragraph{RNN Operation.}
Figure~\ref{fig:rnnpipeline} shows the layout of the sequence of operation in the RNN. Each pixel contains 6 analog registers, named A,B,C,D,E,and F. We refer to register plane A as all the A registers across the entire image sensor. In Figure~\ref{fig:rnnpipeline}, the $256\times256$ pixels are split into a $4\times4$ larger processor elements of size $64\times64$. In register plane A, we take the output from the CNN and the previous hidden state and duplicate it to two other PEs in plane A. Register plane B holds the corresponding weights $\textbf{w}_f$, $\textbf{u}_f$, $\textbf{w}_o$, $\textbf{u}_o$ for the convolution operators needed. 4 convolutions are simultaneously run on one register plane. The outputs in plane B are shifted and added. A binarization is applied to get $\textbf{f}_t$. This is then used to update a hidden state via element-wise multiplication every time step. Every 16 time steps, SCAMP-5 outputs the $64 \times 64$ image corresponding to the output gate $\textbf{o}_t$. Our spatio-temporal feature encoder distills the salient information while giving a 64$\times$ decrease in bandwidth. 
{\paragraph{Accounting for Analog Uncertainty.}
As with all analog compute, a certain amount of noise should be expected. As such, we treat each of the SCAMP's analog registers to contain values split across equally spaced discrete intervals. During convolutions, the binary image and binary weights are XNOR-ed. Depending on the result, we either add or subtract an analog value approximately equal to $10$. As the analog registers have an approximate range of values $-128$ to $127$, the interval cannot be increased without risking saturation during convolutions.  However, there is uncertainty when it comes to the precision and uniformity of the intervals. Along with decay, this uncertainty, spatial non-uniformity, and noise affect the operations that follow. In the RNN, these effects accumulate for 16 frames, leading to a significant amount noise. To account for these effects, we trained binary models in simulation with varying amounts of added Gaussian noise in the CNN and the RNN prior to quantization of the features. We also fine-tuned the off-sensor layer on the training set outputs from SCAMP-5. 
\paragraph{Assessment.}
To test our prototype, we uploaded the video datasets to SCAMP-5 in sequence and saved out the outputs every 16 frames (see supplement for additional details). In our current implementation, it takes roughly 95 ms to run a single frame through the on-sensor encoder. The $64\times64$ output region then goes through the off-sensor linear layer decoder.  We evaluate the performance using the models trained with and without noise. The results shown in Table~\ref{tab:scamp_table} highlight the benefits of training with noise, as well as the difficulty that comes with working with analog registers. We see even running the same train set through SCAMP-5 two separate times does not result in the same performance. Without the noise-trained model, we reached 61.1\% on the hand gesture recognition test set. Performance improved to 73.3\% when we used the weights from training with noise. Similarly, the performance on lip reading was boosted to 70.0\% when using a model trained on noise. While added noise during training helps, the noise characteristics of SCAMP-5 are much more complex.  
Such issues may be mitigated in future sensor–processors with sufficient digital registers to avoid having to rely upon the use of analog computation.
While limited by noise, we demonstrated the first in-pixel spatio-temporal encoder for bandwidth compression. 

\begin{table}[]
    \centering
    \scriptsize
    \begin{tabular}{cccc}
    \toprule
            & Train Set & Train Set & Test Set\\
            & Accuracy run 1 & Accuracy run 2 & Accuracy  \\
         \midrule
         \textbf{Hand Gesture Recognition} & & &\\
         Noise-free Model & 100.0\% & 64.0\% & 61.1\% \\
         Model trained with noise & 95.3\% & 70.8\% & 73.3\% \\
         \midrule
         \textbf{Lip Reading} & & &\\
         Noise-free Model & 100.0\% & 78.9\% & 50.0\% \\
         Model trained with noise & 98.5\% & 84.85\% & 70.0\% \\
        \bottomrule
    \end{tabular}
    \caption{{\bf Experimental Results.} We run the training sets through the SCAMP-5 implementation twice. The first outputs are used to fine-tune the off-sensor linear layer decoder. In theory, the train set accuracy of the second runs should be close. With the noise accumulated through the analog compute, however, the SCAMP-5 implementation is not deterministic. Adding Gaussian noise during training increases the test-set performance. }
    \label{tab:scamp_table}
\end{table}

\section{Discussion}
\label{sec:discussion}
In the traditional computer vision pipeline, full-frame images are extracted from the camera and are fed into machine learning models for different perception tasks. While completely viable in systems not limited by compute, memory, or power, many edge devices do not offer this luxury. For systems like AR/VR devices, robotics, and wearables, low-power operation is crucial, and even more so if the system requires multiple cameras. The community has already been working on creating smaller, more efficient models, as well as specialized accelerators. However, the communication between camera and processor that consumes nearly 25\% of power in these systems \cite{GomezPower} has not been optimized. In this work, we demonstrate how running a simple in-pixel spatio-temporal feature extractor can decrease the bandwidth, and hence power associated with readout, by 64$\times$. Even with highly quantized weights and signals and a very simple decoder, we still maintain good performance on hand gesture recognition and lip reading. We studied different RNN architectures and presented PixelRNN that performs well in highly quantized settings, we studied just how small we could make the bandwidth before affecting performance, shown in Figure~\ref{fig:bandwidth_study}, and implemented a physical prototype with one of the emerging sensors, SCAMP-5, that is paving the way for future sensors. 

\paragraph{Limitations and Future Work.}
{One of the biggest challenges of working with SCAMP5 is accounting for the analog noise, but the platform offers great flexibility to program the data movement between pixels to implement prototypes.  While SCAMP-5 offers many exciting capabilities, it is still limited in memory and compute as all circuitry needs to fit in a single pixel. Until recently, adding circuitry or memory to image sensors compromised the fill factor, which worsens the imaging performance and limits achievable image resolution. With the new developments in stacked CMOS image sensors, future sensors will be able to host much more compute and memory on the sensor plane, allowing us to design more expressive models and to apply tools like architecture search to optimize where compute happens in a system \cite{Splitnets}. Until then, we are limited to light-weight in-pixel models. Noise related to analog compute can also be mitigated by switching over to digital compute. Our work is not only applicable to SCAMP-5 but to all future focal-plane processors. }

\paragraph{Conclusion.} Emerging image sensors offer programmability and compute directly in the pixels. Our work is the first to demonstrate how to capitalize on these capabilities using efficient RNN architectures, decreasing the bandwidth of data that needs to be read off the sensor as well as stored and processed by downstream application processors by a factor of 64$\times$. We believe our work paves the way for other inference tasks of future artificial intelligence--driven sensor--processors.

\section*{Acknowledgements}
This project was in part supported by the National Science Foundation and Samsung.

{\small
\bibliography{main.bbl}
}

\newpage

\setcounter{figure}{0}
\setcounter{section}{0}
\setcounter{equation}{0}

\clearpage
\section*{\Large Supplemental Material}
\renewcommand{\thefigure}{S\arabic{figure}}
\renewcommand{\thesection}{S\arabic{section}}
\renewcommand{\thetable}{S\arabic{table}}

\section{Additional Details on Comparisons}

\subsection{Baseline Architectures}

In this section, we discuss additional details of the baseline approaches and network architectures considered in the paper. For each baseline, we list the equations modeling the on-sensor computation. The output of this step is read off the sensor, transmitted to a host processor, and processed by another full-precision fully-connected layer there. This fully-connected layer is trained end to end for each baseline, but omitted in the following equations. The size of the weight matrix of this layer is $N \times M$, where $N$ is the number of elements of the output of a baseline (i.e., it varies per baseline) and $M$ is the number of categories to classify in the respective dataset ($M=9$ for hand gesture recognition and $M=4$ for lip reading). Note that this off-sensor model can further be optimized for a given task, but here we use the simplest off-sensor model to demonstrate the benefits of our spatio-temporal feature encoder.

{\bf Source code for all baselines will be made available.}

Table~1 shows an overview of all baselines, including their Top-1 accuracy for the two tasks, the number of parameters in the feature encoder, the memory required to store intermediate features in the forward pass, and the total required bandwidth.

\subsubsection{RAW Mode}

The RAW camera mode simply outputs the input with a resolution of $256 \times 256$ pixel at 8~bit precision at every time step. This data is sent off the sensor and processed by a fully-connected layer there. The output layer on the sensor, $\textbf{o}_t$, is simply the input $\textbf{x}_t$ at each time step $t$:
\begin{align}
    \textbf{o}_t = \textbf{x}_t.
\end{align}
The accuracy of this model is calculated as the number of frames classified correctly out of the total 16 frames per video. 

\subsubsection{Difference Camera}
The difference camera is simple temporal feature encoder. 
\begin{align}
    \textbf{h}_t &= \textbf{x}_t \\
    \textbf{o}_t &= \psi_o \left ( \textbf{x}_t - \textbf{h}_{t-1}   \right )
\end{align}
where 
$\psi_o \left(x \right) = \left\{ \footnotesize{\begin{array}{rl}
        -1 & \text{for } \textrm{x} < - \delta\\
        0 & \text{for } -\delta \leq \textrm{x} \leq \delta\\
        1 & \text{for } \delta < \textrm{x}
        \end{array} } \right. ,
$ for some threshold $\delta$. The difference camera output is sparse. For comparison, we output $64 \times 64$ ternary frame indicating the polarity of the pixel at every time step $t$.  For this purpose, we add a hidden state $\textbf{h}_{t-1}$ that ``remembers'' the last intensity value of each pixel. The output at each frame is then the difference between the current intensity value at a pixel and the hidden state, i.e., the intensity recorded at the last frame.

\subsubsection{Recurrent Neural Network (RNN) Architectures}

We next describe several RNN architectures that we have tested at full precision and also using binary weights. We list the equations that readers will be most familiar with in the following. Note that we do not use bias values for any of these RNNs, as such biases are challenging to incorporate into our current PPA hardware platform. Moreover, in the binary setting, all nonlinear activations  functions $\psi$ are replaced by the $\textrm{sign}$ function in the forward pass and the gradient of the $\textrm{tanh}$ function for backpropagation during training (this is smoother and more robust than naively binarizing the weights at the end of training or using the straight-through-estimator). In all cases $\text{CNN} \left( \cdot \right)$ refers to a CNN-based feature encoder operating directly on the intensity, $\textbf{x}_t$, at time step $t$. Unless otherwise noted, the output is read from the sensor only every 16 time steps and then processed off-sensor by the aforementioned fully-connected layer. The number of values output from each RNN is fixed to 4096 values every 16 time steps, as this is the number our hardware platform can achieve in practice. In the main paper, we also show how decreasing bandwidth deteriorates accuracy.

\paragraph{Simple Convolutional RNN (SRNN).}

The SRNN baseline is the simplest RNN that uses a hidden state $\mathbf{h}_t$ at each time step as
\begin{align}
    \mathbf{\tilde{h}}_t &=  \mathbf{w}_h \! * {\text{CNN} \left(\mathbf{x}_t\right) } + \mathbf{u}_h*\mathbf{h}_{t-1} + \textbf{b}_h,\\
    \mathbf{h}_t &=  \psi_h ( \mathbf{\tilde{h}}_t ),\\
     \mathbf{o}_t &= \mathbf{\tilde{h}}_t
\end{align} where $ \psi_h=\text{tanh}()$ function. As mentioned before, we set bias values $\textbf{b}_h$ for this architecture and all following to 0 to be meaningful for our hardware platform.

\paragraph{Convolutional Long Short-term Memory (LSTM)}

Long short-term memory (LSTM) models~\cite{LSTM} are among the most well-known RNN architecture. Originally introduced to mitigate the vanishing gradient problem during RNN training, an LSTM usually provides a hidden state $\mathbf{h}_t$, cell state $\mathbf{c}_t$, an input $\mathbf{i}_t$, output $\mathbf{o}_t$, and forget $\mathbf{f}_t$ gates. We use a convolutional variant of an LSTM:
\begin{align}
    \mathbf{f}_t &= \psi_f \left ( \mathbf{w}_f \! * {\text{CNN} \left(\mathbf{x}_t\right) }+ \mathbf{u}_f*\mathbf{h}_{t-1} +\textbf{b}_f \right ),\\
    \mathbf{i}_t &= \psi_i \left ( \mathbf{w}_i \! * {\text{CNN} \left(\mathbf{x}_t\right) }+ \mathbf{u}_i*\mathbf{h}_{t-1} +\textbf{b}_i \right ),\\
    \mathbf{o}_t &=\psi_o \left ( \mathbf{w}_o \! * {\text{CNN} \left(\mathbf{x}_t\right) }+ \mathbf{u}_o*\mathbf{h}_{t-1} +\textbf{b}_o \right ),\\
    \mathbf{\tilde{c}}_t &= \psi_{\tilde{c}} \left ( \mathbf{w}_{\tilde{c}} \! * {\text{CNN} \left(\mathbf{x}_t\right) }+ \mathbf{u}_{\tilde{c}}*\mathbf{h}_{t-1}  +\textbf{b}_{\tilde{c}}\right ),\\
    \mathbf{c}_t &= \mathbf{f}_t \odot \mathbf{c}_{t-1}+ \mathbf{i}_t \odot  \mathbf{\tilde{c}}_t,\\
    \mathbf{h}_t &= \mathbf{o}_t \odot \sigma (\mathbf{c}_{t})
\end{align} where $ \psi_f= \psi_i = \psi_o = \psi_{\tilde{c}} = \sigma()$, the sigmoid activation function. Note that the LSTM in its standard notation has an output gate $\mathbf{o}_t$, but (confusingly), this is not what is actually read off the sensor. For the LSTM, the hidden state $\mathbf{h}_t$ is actually read off the sensor and further processed by the full-connected layer.

\paragraph{Convolutional Gated Recurrent Unit (GRU).}

The GRU~\cite{GRU} is among the most popular RNN architectures. It provides two gates: an update gate $\textbf{z}_t$ and a reset gate $\textbf{r}_t$. GRUs have been introduced as leaner variants of LSTM that offer similar performance in many application with fewer parameters. The convolutional GRU we use is 
\begin{align}
    \mathbf{z}_t &= \psi_z \left ( {\mathbf{w}_z *}{\text{CNN} \left(\mathbf{x}_t\right) } + \mathbf{u}_z*\mathbf{h}_{t-1} + \mathbf{b}_z \right ),\\
    \mathbf{r}_t &= \psi_r \left ( {\mathbf{w}_r *}{\text{CNN} \left(\mathbf{x}_t\right) } + \mathbf{u}_r*\mathbf{h}_{t-1} + \mathbf{b}_r \right ),\\
    \tilde{\mathbf{h}}_t &= \psi_{\tilde{h}}\left (\mathbf{w}_h \!\! * \!{\text{CNN} \left(\mathbf{x}_t\right) } \! + \! \mathbf{u}_h \!\! * \! (\mathbf{r}_t \odot \mathbf{h}_{t-1}) \! + \! \mathbf{b}_h  \right ),\\
     \mathbf{h}_t &= (1-\mathbf{z}_t) \odot \mathbf{h}_{t-1} + \mathbf{z}_t \odot \tilde{\mathbf{h}}_t,
  \mathbf{o}_t = \mathbf{h}_t ,
\end{align}  where $\psi_z=\psi_r=\sigma$ and $\psi_{\tilde{h}}=\text{tanh}()$.

Similar to the LSTM, for a GRU the hidden state $\mathbf{h}_t$ is read off the sensor and further processed by the full-connected layer. There is not output gate in this case.

\paragraph{Convolutional Minimal Gated Unit (MGU).}
MGU~\cite{minimalGRU} were introduced as the minimal variant of gated units, i.e., they only use a single gate which is treated as the forget gate $\mathbf{f}_t$. The convolutional variant we use is
\begin{align}
    \mathbf{f}_t &= \psi_f \left ( {\mathbf{w}_f *}{\text{CNN} \left(\mathbf{x}_t\right) } + \mathbf{u}_f*\mathbf{h}_{t-1} + \mathbf{b}_f \right ),\\
    \tilde{\mathbf{h}}_t &= \psi_{\tilde{h}}\left (\mathbf{w}_h \!\! * \! {\text{CNN} \left(\mathbf{x}_t\right) } \! + \! \mathbf{u}_h*(\mathbf{f}_t \odot \mathbf{h}_{t-1}) \! + \! \mathbf{b}_h  \right ),\\
     \mathbf{h}_t &= (1-\mathbf{f}_t) \odot \mathbf{h}_{t-1} + \mathbf{f}_t \odot \tilde{\mathbf{h}}_t,
  \mathbf{o}_t = \mathbf{h}_t ,
\end{align} where $\psi_f = \sigma() $ and $\psi_{\tilde{h}}= \text{tanh}()$.

Again, the hidden state $\mathbf{h}_t$ is read off the sensor and further processed by the full-connected layer.

\paragraph{{Developing the PixelRNN Architecture}}
We start with a simple, intuitive formulation of an RNN, inspired by the GRU and MGU. We call this the general RNN (\textbf{genRNN})
\begin{align}
    \textbf{f}_t &= \psi_f \left( \textbf{w}_f* \textbf{CNN}(\text{x}_t) + \textbf{u}_f * \textbf{h}_{t-1} + \textbf{b}_f\right)\\
     \tilde{\textbf{h}}_t &= \psi_{\tilde{h}} \left( \textbf{w}_h* \textbf{CNN}(\text{x}_t) + \textbf{u}_h * \textbf{h}_{t-1} +\textbf{b}_{\tilde{h}} \right)\\
     \mathbf{h}_t &= (1-\mathbf{f}_t) \odot \mathbf{h}_{t-1} + \mathbf{f}_t \odot \tilde{\mathbf{h}}_t \\
      \textbf{o}_t &= \psi_o \left( \textbf{w}_o* \textbf{CNN}(\text{x}_t) + \textbf{u}_o * \textbf{h}_{t-1} + \textbf{b}_o\right)
\end{align} where $\psi_f= \psi_{\tilde{h}} = \sigma()$ and $\psi_o$ an optional non-linear activation.  The hidden state is updated by interpolating between the previous hidden state $\textbf{h}_{t-1}$ and the new candidate hidden state $\tilde{\textbf{h}}$.
In the highly quantized binary setting, we simplify these equations even more by setting
\begin{align}
    \tilde{\textbf{h}}_t = -\textbf{h}_{t-1}
\end{align} which is computationally simple and proved to work well in practice when the hidden state is re-initialized to all ones every $K$ frames and the weights are either -1 or 1. We can simplify the genRNN form and arrive at \textbf{PixelRNN}'s architecture:
\begin{align}
    \mathbf{f}_t &= \psi_f \left ( \mathbf{w}_f \! * {\text{CNN} \left(\mathbf{x}_t\right) }+ \mathbf{u}_f*\mathbf{h}_{t-1} \right ),\\
    \mathbf{h}_t &= \mathbf{f}_t \odot \mathbf{h}_{t-1},\\
    \mathbf{o}_t &= \psi_o \left (\mathbf{w}_o * {\text{CNN} \left(\mathbf{x}_t\right) } + \mathbf{u}_o * \mathbf{h}_{t-1}  \right ).
\end{align}

\subsubsection{Synchronous Event Camera Approximation}
As a temporal-only encoder, we look to the event camera. The ideal event camera is asynchronous and would call for a completely different kind of neural network architecture. In practice, events are time-stamped, effectively creating a synchronous event camera. One of the advantages of event cameras is that the readouts are often sparse if most of the scene is not changing. Here, we approximate a synchronous event camera that sends out a $64 \times64$ binary image encoding the locations of events each timestep. The genRNN architecture can approximate this when the CNN layer is omitted, i.e. CNN() is an identity function, $\textbf{w}_f=\textbf{w}_h=\textbf{w}_o = 1$, $\textbf{u}_f=\textbf{u}_o=-1$, $\textbf{u}_h=0$, $\psi_h$ is the identity function. This leads to
\begin{align}
    \textbf{f}_t &= \psi_f(\text{x}_t - \textbf{h}_{t-1})\\
    \textbf{h}_t &=  \textbf{f}_t \odot (\text{x}_t-\textbf{h}_{t-1})+ (1-\textbf{f}_t) \odot \textbf{h}_{t-1}\\
    \textbf{o}_t &= \psi_o (\text{x}_t - \textbf{h}_{t-1})
\end{align}
where $\psi_f \left(x \right) = \left\{ \footnotesize{\begin{array}{rl}
        1 & \text{for } |\textrm{x}| > \delta\\
        0 & \text{for } |\textrm{x}| \leq \delta
        \end{array} } \right.\\$ and 
   $ \psi_o \left(x \right) = \left\{ \footnotesize{\begin{array}{rl}
       1 & \text{for } x > \delta\\
        0 & \text{for } |\textrm{x}| \leq \delta\\
        -1 & \text{for } x < \delta
        \end{array} } \right.$\\
We include the baseline comparison here in the supplement as an additional temporal-only encoder. The readout bandwidth is calculated as $64 \times 64 $ `events' (values are -1 or 1). Unlike the RNNs, the event camera approximation does not compress temporally, so we evaluate the accuracy by averaging the scores over the 16 frames and use the largest as the prediction for the video.

\begin{figure*}[t!]
    \centering
    \includegraphics[clip, trim=0cm 0cm 1cm 0cm, width=\columnwidth]{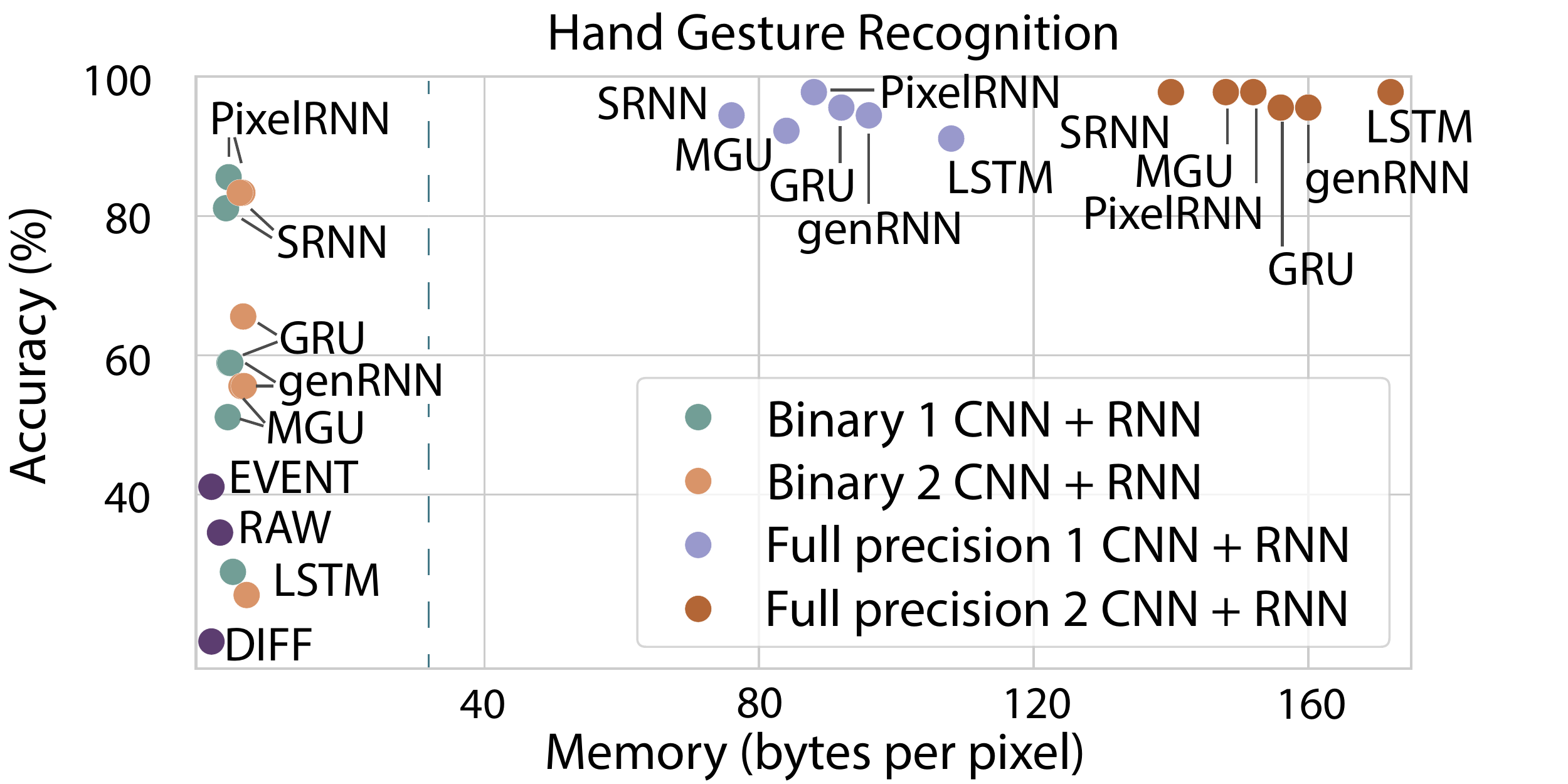}
    \hspace{1em}
    \includegraphics[clip, trim=0.5cm 0cm 1cm 0cm, width=\columnwidth]{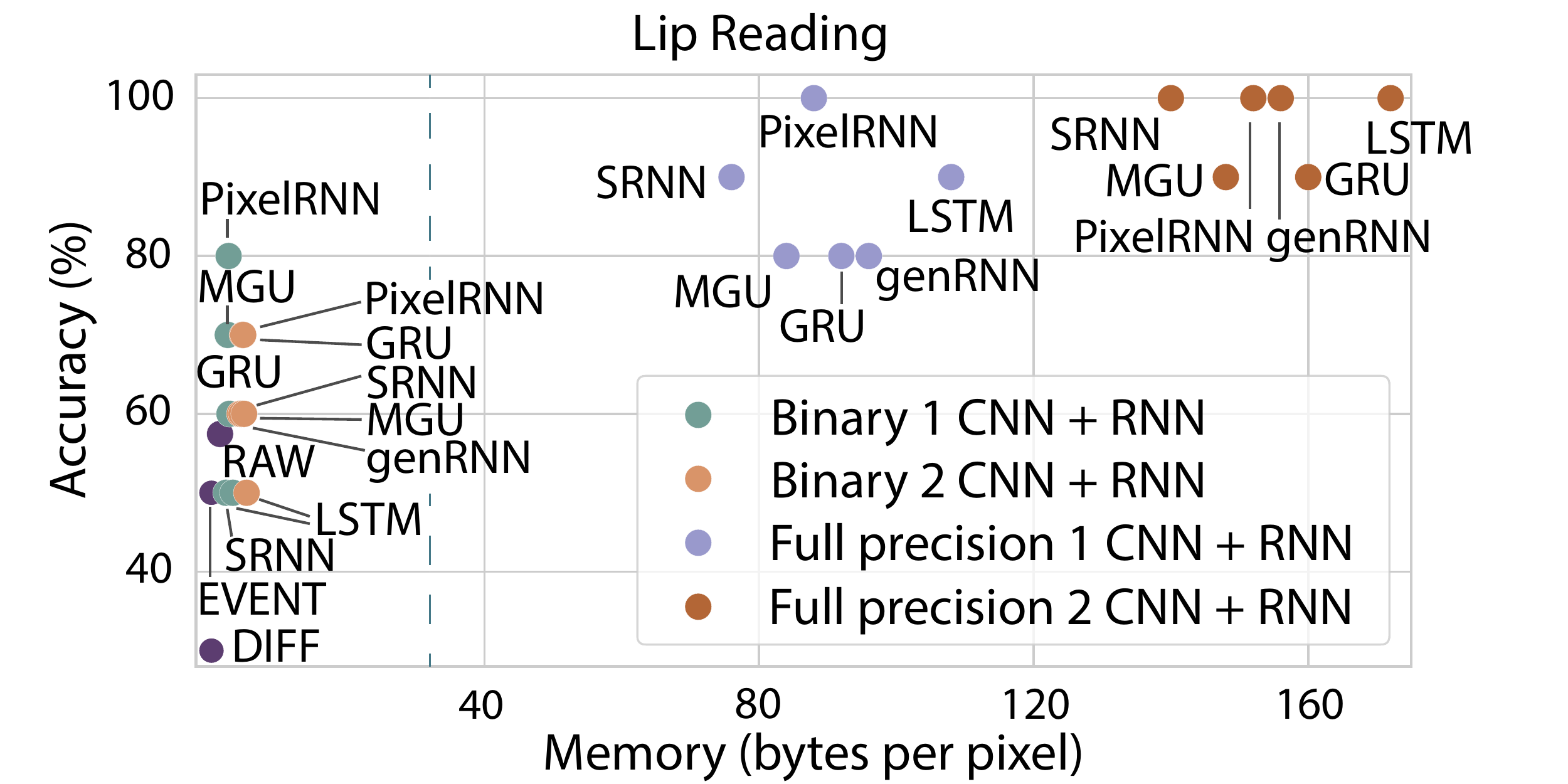}
    \caption{{\bf Network architecture comparison.} We plot the accuracy of several baseline architectures on two tasks: hand gesture recognition (left) and lip reading (right) on the test sets of the Cambridge Hand Gesture dataset and the Tulips1 dataset, respectively. As baselines, we use RAW, difference camera, approximate event camera, as well as the following RNN architecture: SRNN, MGU, GRU, LSTM, genRNN in (0,1) binary regime and PixelRNN, each with 1- and 2-layer CNN encoders and binary or full 32 bit floating point precision. For the full-precision networks, PixelRNN offers the best performance for the lowest memory footprint. Comparing the networks with binary weights, PixelRNN also offers the best accuracy. Surprisingly, larger architectures, such as GRUs and LSTMs, do not perform well when used with binary weights. This can be explained by the increasing difficulty of reliably training increasingly large networks with binary parameter constraints. Leaner networks, such as PixelRNN, can be trained more robustly and reliably in these settings. Note that the memory plotted on the x-axis represents all intermediate features, per pixel, that need to be stored when running an input image through the RNN. We do not count the network parameters in this plot, because they do not dominate the memory requirements and are shared among all pixels. }
    \label{fig:accuracyplots}
\end{figure*}

\subsection{Additional Experiments}

\paragraph{Binarization Regimes: } As mentioned above, the non-linearites in the RNN architectures become sign functions when operating in the $-1,1$ binary regime, unless otherwise noted. The hidden states are similarly binarized. Alternatively, we could operate in the $0$ and $1$ regime. These lead to slightly different interpretations. In the $0,1$ case, information can be lost when it is set to 0. In the $-1,1$ case however, no information is ``lost.'' Rather the values in the hidden state will flip between $-1$ and $1$. To determine which regime to use, we tested all the architectures with both binarizations. To train the models effectively, we use the following gradient estimators.

In the $-1,1$ case, we utilize the gradient of $\tanh(mx)$ as a gradient estimator for the sign function
 \begin{align}
     \text{Forward: }& w = q(\tilde{w}) =\text{sign}(\tilde{w})\\
     \text{Backward: }& \frac{\partial q}{\partial \tilde{w}} \approx m \cdot (1-\tanh^2(m\tilde{w}))
 \end{align}
  where $m$ can be used to tune how steep the transition between $-1$ and $1$ is. We test both our PixelRNN architecture as well the general architecture outlined in equations 20-23.
  
 In the $0,1$ case, we utilize the analogous gradient of the sigmoid function $\frac{1}{1+e^{-mx}}$ to estimate the gradient
 \begin{align}
    \text{Forward: }& w = q(\tilde{w}) =\psi(\tilde{w})\\
    \text{Backward: }& \frac{\partial q}{\partial \tilde{w}} \approx \frac{m\cdot e^{-mw}}{(1+e^{-mw})^2}
 \end{align}
where $\psi_f \left(x \right) = \left\{ \footnotesize{\begin{array}{rl}
        1 & \text{for } w > 0\\
        0 & \text{for } w \leq 0
        \end{array} } \right.\\$ and $m$ is used to tune the steepness of the transition between $0$ and $1$. We found binarizing to $-1$ and $1$ had higher performances than when binarizing to $0$ and $1$. 

\paragraph{Hidden State Initialization: } We tested different hidden state initializations, initializing to all zeros or all ones every $K$ frames. In most cases, the initialization did not make a difference. In some cases, however, initializing to all ones performed slightly better on Hand Gesture recognition and Lip reading. For this reason, we initialize the hidden state to all ones every $K$ frames.

\paragraph{Extended Architecture Results: }
Along with the tested architectures shown in the main paper, we include the performances of our starting architecture genRNN and the synchronous event camera approximation in the extended ablation on the same \href{https://labicvl.github.io/ges_db.htm}{Cambridge Hand Gesture Recognition} dataset and the \href{https://inc.ucsd.edu/mplab/36/}{Tulips1} lip reading dataset. Extended results can be seen in Table~\ref{tab:table_supp} and Figure~\ref{fig:accuracyplots} (left). We evaluated all baselines described above on the hand gesture recognition task and (right) lip reading task. The figure shows the the Top-1 accuracy for all baselines in addition to the memory required by intermediate features of the respective network. This memory footprint is normalized per pixel and does not include the network parameters themselves, because these are shared across all pixels and, in the binary case, add a negligible amount of memory. The intermediate features are the bottleneck, especially for higher-resolution sensors using small RNN architectures with binary weights. Each RNN baseline is tested with the same one- or two-layer CNN feature encoder as well as using full 32 bit floating point precision and 1 bit binary features. 

For the full-precision networks, PixelRNN offers the same accuracy as other top-performing RNNs, but with a significantly reduced memory footprint. Comparing the networks with binary weights, PixelRNN also offers the best accuracy. Surprisingly, larger architectures, including GRUs and LSTMs, do not perform well when used with binary weights. This can be explained by the increasing difficulty of reliably training larger RNN networks with binary parameter constraints as the exploding gradient problem worsens \cite{lstm_quantized}. Leaner networks, such as SRNN and PixelRNN, can be trained more robustly and reliably in these settings.

\begin{table*}[]
    \centering
    \small
    \begin{tabular}{|c|c|c|c|c|c|c|}
    \hline
         Model Name & \begin{tabular}{@{}c@{}c@{}}Hand Gesture \\ Top-1 Accuracy \%\\ (1 bit / 32 bit)\end{tabular} &\begin{tabular}{@{}c@{}c@{}}Lip Reading \\ Top-1 Accuracy \% \\ (1 bit / 32 bit)\end{tabular} &  \begin{tabular}{@{}c@{}c@{}}\# Model\\ Params\end{tabular} & \begin{tabular}{@{}c@{}}Feature Memory in\\ bytes per pixel\\ (1 bit / 32 bit) \end{tabular} & \begin{tabular}{@{}c@{}}Readout Bandwidth \\ (values per 16 timesteps)\end{tabular}   \\
         \hline
         RAW                 & \quad\quad--- / 34.53\% & \quad\quad--- / 57.45\% &0& --- / 1.50 &  256 $^2$ $\cdot$ 16 = 1,048,576 \\
         Difference Camera        & 18.89\% / \quad---\quad\quad   &30.00\% / \quad---\quad \quad  &0 & 0.25 /\quad---&  64 $^2$ $\cdot$ 16 = 65,536\\ 
         Event Camera*        & 41.11\% / \quad---\quad\quad   &50.00\% / \quad---\quad \quad  &0 & 0.25 /\quad---& 65,536\\
         \hline
         1 CNN + SRNN  & 81.11\% / 94.44\% & 50.00\% / 90.00\% & 451 &  2.38 / 76.00 &  4,096  \\
         1 CNN + LSTM  & 28.89\% / 91.11\% & 50.00\% / 90.00\%  & 601 &  3.38 / 108.00 & 4,096 \\
         1 CNN + GRU  & 58.89\%  / 95.56\% & 60.00\% / 80.00\% & 551 &  2.88 / 92.00&  4,096  \\
         1 CNN + MGU  & 51.11\% / 92.22\% & 70.00\% / 80.00\% & 501 &  2.63 / 84.00 &  4,096  \\
         1 CNN + genRNN (0,1)  & 58.89\% / 94.44\% & 60.00\% / 80.00\% & 553 & 3.00 / 96.00  &  4,096  \\
         1 CNN + genRNN (-1,1)  & 84.44\% / 94.44\% & 80.00\% / 80.00\% & 553 & 3.00 / 96.00  &  4,096  \\
         1 CNN + PixelRNN & \textbf{85.56\% }/ \textbf{97.78\% }& \textbf{80.00\% / 100.00\% }& 501 & 2.75 / 88.00 &   4,096  \\ \hline
         2 CNN + SRNN  & 83.33\% / \textbf{97.78\% } & 60.00\% / \textbf{100.00\%} & 852 &  4.38 / 140.00 &  4,096  \\
         2 CNN + LSTM  & 25.56\% / \textbf{97.78\% } & 50.00\% / \textbf{100.00\%}& 1,002 &  5.38 / 172.00 &  4,096  \\
         2 CNN +GRU  & 65.56\% / 95.56\% & 70.00\% / \textbf{100.00\%} & 952 &  4.88 / 156.00 &  4,096  \\
         2 CNN +MGU  & 55.56\% / \textbf{97.78\%} & 60.00\% / 90.00\%  & 902 &  4.63 / 148.00 &  4,096  \\
         2 CNN + genRNN (0,1)  & 55.56\% / 95.56\% & 60.00\% / 90.00\% & 954 &  5.00 / 160.00 &  4,096  \\
         2 CNN + genRNN (-1,1)  & 83.33\% / 95.56\% & 60.00\% / 90.00\% & 954 & 5.00 / 160.00  &  4,096  \\
         2 CNN + PixelRNN & \textbf{87.78\%} / \textbf{97.78\% } & \textbf{70.00\%} / \textbf{100.00\%} & 902 & 4.75 / 152.00 &   4,096  \\\hline
    \end{tabular}
    \caption{{\bf Network architecture comparison.} We evaluate the accuracy of several baseline architectures on two tasks: hand gesture recognition and lip reading. As baselines, we use RAW and event cameras as well as the following RNN architecture: SRNN, MGU, GRU, LSTM, genRNN, and PixelRNN, each with 1- and 2-layer CNN encoders and binary or full 32 bit floating point precision. as PixelRNN has better accuracy than the genRNN and fewer operations and memory required, we chose PixelRNN to be our final architecture. In this table, we also list the number of parameters for each model, which includes the CNN feature encoder and the respective RNN. The feature memory column lists the memory footprint of all intermediate features that need to be computed during a forward pass through each network. This does not include the model weights. Finally, the readout bandwidth lists the number of values that are read out in total within a 16 frame sequence.  } 
    \label{tab:table_supp}
\end{table*}

\section{{Additional Details}}
\begin{figure*}[t]
    \centering
    \includegraphics[clip, trim=0cm 13cm 0.5cm 0cm, width=1\linewidth]{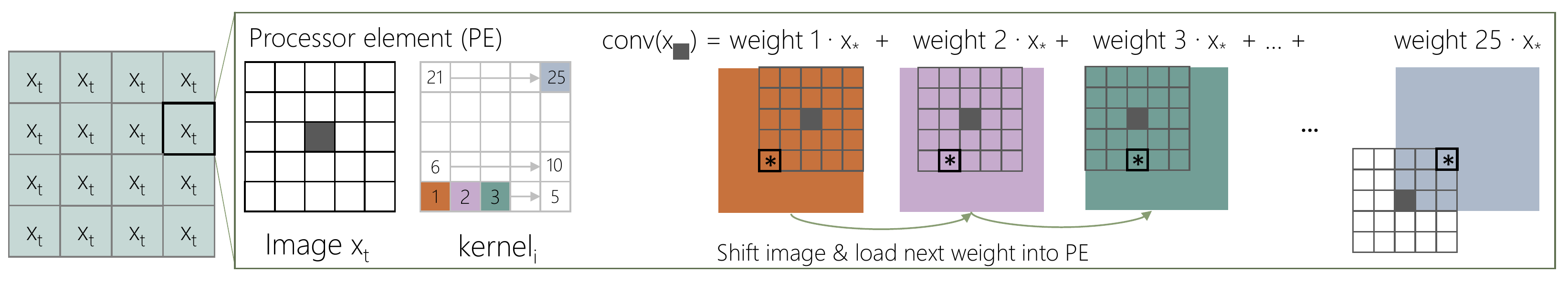}
    \caption{{The CNN and the gates in our RNN are convolutional. Above is a block diagram of the data movement in SCAMP-5 to calculate a convolution. The $256 \times 256$ image plane is split into a $4 \times 4$ grid, with each space containing $64 \times 64$ processor elements (PE), each performs a parallel convolution between $x_t$ and a kernel. In a convolution, the first weight in the $5 \times 5$ kernel multiplies with the image, and the result is added to a running sum in the pixel. The image is then shifted and the next weight is loaded, repeating this process for all 25 weights. See Bose et al.~\cite{boseECCV2020} for additional details. } 
    }
    \label{fig:convolution}
\end{figure*}
\begin{figure*}
    \centering
    \includegraphics[clip, trim=0cm 4cm 3cm 0cm, width=0.95\linewidth]{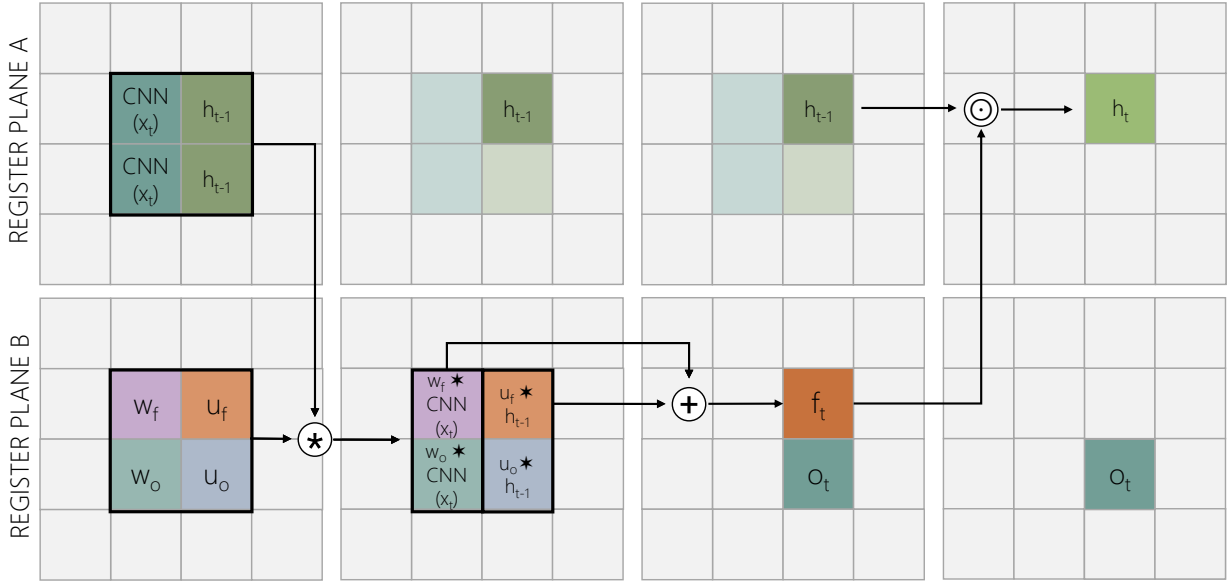}
    \caption{{To implement the PixelRNN on SCAMP-5, the image plane is split into a $4 \times 4$ grid shown above, each grid space comprising a $64 \times 64$ block of processor elements. We make use of two analog register planes, A and B, in the four central PEs blocks. Above, we show the sequence of operations from left to right. The input from the CNN and the previous hidden state are duplicated in A. These 4 blocks of $64 \times 64$ PEs are convolved $*$ with the corresponding gate weights stored in plane B. The resulting convolutions in the second column are then added to compute the output $\textbf{o}_t$ and the forget gate $\textbf{f}_t$. Note that an in-place binarization is applied to $\textbf{f}_t$. The hidden state $\textbf{h}_t$ is updated via an element-wise multiplication $\odot$  of $\textbf{h}_{t-1}$ and $\textbf{f}_t$ } 
    }
    \label{fig:rnn_op}
\end{figure*}

\subsection{Additional Implementation Details}
\paragraph{Training Details: } The Cambridge hand gesture dataset has 900 videos while the Tulips dataset has 96 videos. In both cases, we use a 80:10:10 train:validation:test split. As they are small datasets, we utilize data augmentation techniques. For each video, we select 16 evenly spaced frames starting with the first frame. Images are resized to $256 \times 256$ for the raw camera and $64 \times 64$ otherwise. During training, an optional, random temporal offset is also applied to the train set, if available. For videos with less than 16 frames, we randomly duplicate a subset of the frames until 16 frames are available. We also apply a random spatial offset to the training videos for further augmentation. During training, we use the Pytorch Adam optimizer and a cross entropy loss. All models were trained for 300 epochs, or until convergence, with learning rate sweeps from $1e-1$ to $1e-4$ and a learning rate scheduler that reduces when the loss and accuracy improvement have plateaued. Best scores for all models were reported.

\subsection{Evaluation Protocol for SCAMP-5 Experiments}
We retrain PixelRNN with Gaussian noise added prior to quantization of the signal everywhere in the model. We then save the trained binary model weights into two of the register planes in SCAMP-5. Weight storage can further be optimized in the future. The 16 image sequence for each test video is extracted from the same pre-processing as in simulation and uploaded sequentially to the PixelRNN. The image is thresholded according to a learned parameter in simulation and subsequently binarized. Due to the architecture of SCAMP-5, we split the sensor into a $4\times4$ grid, comprising $64\times64$ blocks of processing elements (PE). The CNN operation that simultaneously computes 16 convolutions in these PEs is illustrated in figure \ref{fig:convolution}. As our weights and signal are $-1$ and $1$, but are stored in a single digital register where $0$ means $-1$, the `multiplying' operation in the binary convolution becomes a XNOR operation, as opposed to the AND operation. Where the weight XNOR the signal = 1, we add an increment of 10 to the running sum in an analog register. Where the weight XOR the signal = 0, we subtract the increment of 10. This is then thresholded with a learned parameter during training and binarized.  This process is used for the CNN as well as for the convolutions in the RNN. 

The RNN operation is shown here again in figure \ref{fig:rnn_op}. The RNN uses just 4 of the 16 available PEs as just 4 convolutions are needed for the gate computations. At the beginning of each video sequence, the hidden state is initialized to all ones, and then the hidden state is updated and binarized each time step. The outputs for each video, i.e. the encoded images are saved out every 16 frames. We perform this process on the train set and use this to fine-tune the off-sensor model, which is in this case a fully connected layer. The test set encoded images are fed into the fully connected layer and the class is predicted. 

\subsection{Noise Characterization of SCAMP-5}

\begin{figure}
    \centering
    \includegraphics[clip, trim=0cm 0cm 0cm 0cm, width=\columnwidth]{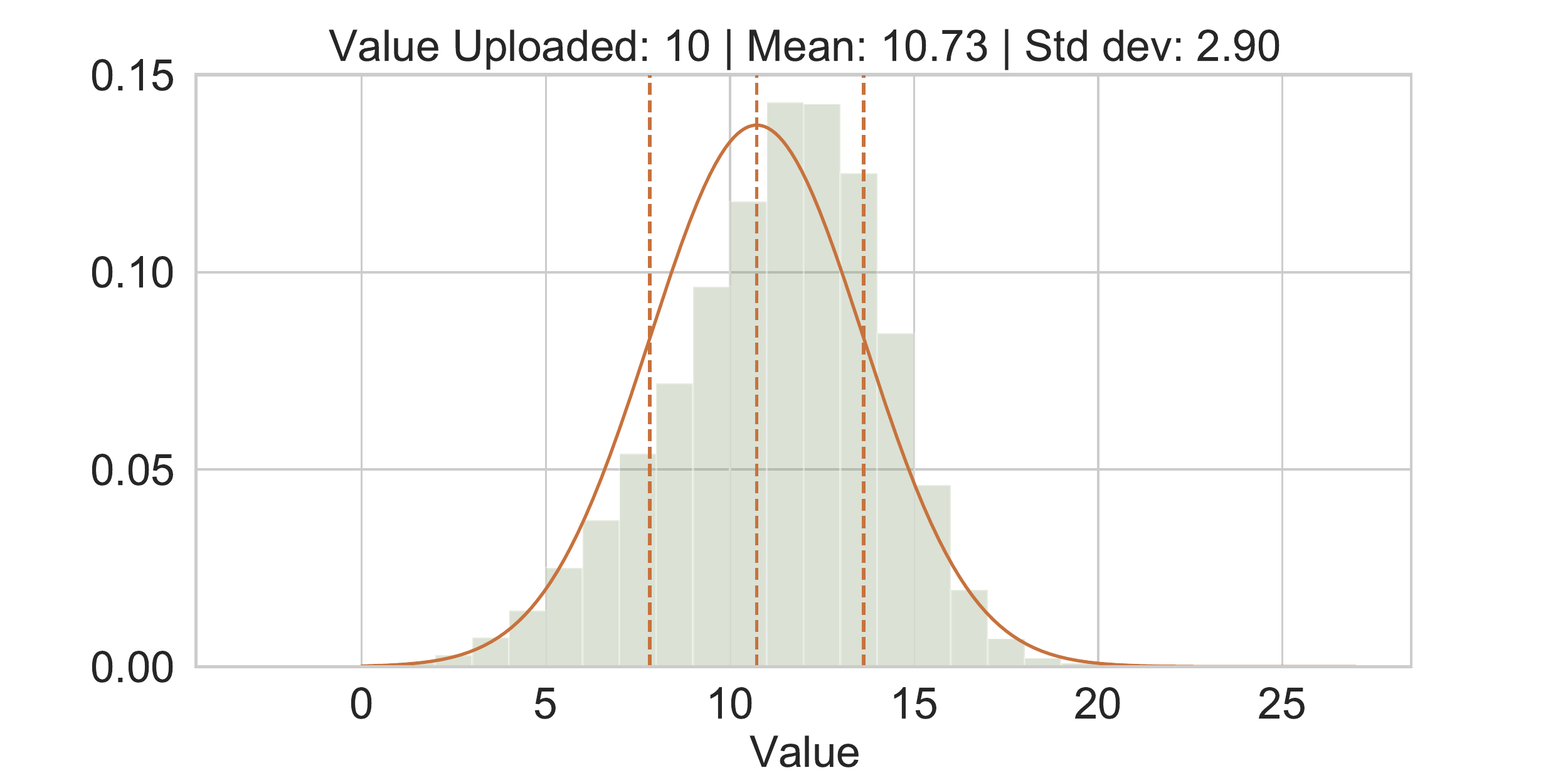}
    \caption{{We estimate the mean and standard deviation of Gaussian noise to add by uploading value 10 to SCAMP-5 and saving out the $256\times256$ image 1,000 times. Above are all the values plotted in a histogram, normalized so the total area is 1. We extract the mean and standard deviation and use these statistics to choose the noise level in simulations.  } 
    }
    \label{fig:hist}
\end{figure}

To decide how much noise to add in training, we send value $10$, i.e. what we use for the intervals in calculating the convolutions, 1,000 times through SCAMP-5, saving out the full $256 \times 256$ image each time. We plot a histogram of the recorded values in fig. \ref{fig:hist}. The histogram is normalized to have an area equal to 1. Overlaid is the normal distribution with the extracted mean $10.73$ and standard deviation $2.90$. We add Gaussian noise with the same standard deviation prior to quantization when training our model for the SCAMP-5 prototype. While the noise illustrated by this plot may be composed of multiple factors including the uncertainty in the recorded values as well as the noise from readout, our first order Gaussian noise approximation improves the model performance on our prototype.

\end{document}